\documentclass[10pt,twocolumn,letterpaper]{article}

\usepackage{cvpr}              

\usepackage[accsupp]{axessibility}  
\usepackage{graphicx}
\usepackage{epstopdf}
\usepackage{amsmath}
\usepackage{amssymb}
\usepackage{booktabs}
\usepackage{pifont}
\usepackage{caption}
\usepackage{float}
\usepackage[pagebackref,breaklinks,colorlinks]{hyperref}

\usepackage[capitalize]{cleveref}
\crefname{section}{Sec.}{Secs.}
\Crefname{section}{Section}{Sections}
\Crefname{table}{Table}{Tables}
\crefname{table}{Tab.}{Tabs.}

\def\cvprPaperID{11790} 
\def\confName{CVPR}
\def\confYear{2023}

\begin{document}

\title{Attribute-preserving Face Dataset Anonymization via Latent Code Optimization}

\author{
Simone Barattin$^{*1}$ \and 
Christos Tzelepis$^{*2}$ \and 
Ioannis Patras$^2$ \and 
Nicu Sebe$^1$ \\ \and
$^1$University of Trento\\
{\tt\small simone.barattin@studenti.unitn.it, niculae.sebe@unitn.it}
\and
$^2$Queen Mary University of London\\
{\tt\small \{c.tzelepis, i.patras\}@qmul.ac.uk}
}

\maketitle
\def\thefootnote{*}\footnotetext{Equal contribution.}

\begin{abstract}
This work addresses the problem of anonymizing the identity of faces in a dataset of images, such that the privacy of those depicted is not violated, while at the same time the dataset is useful for downstream task such as for training machine learning models. To the best of our knowledge, we are the first to explicitly address this issue and deal with two major drawbacks of the existing state-of-the-art approaches, namely that they (i) require the costly training of additional, purpose-trained neural networks, and/or (ii) fail to retain the facial attributes of the original images in the anonymized counterparts, the preservation of which is of paramount importance for their use in downstream tasks. We accordingly present a task-agnostic anonymization procedure that directly optimizes the images' latent representation in the latent space of a \textit{pre-trained} GAN. By optimizing the latent codes directly, we ensure both that the identity is of a desired distance away from the original (with an identity obfuscation loss), whilst preserving the facial attributes (using a novel feature-matching loss in FaRL's~\cite{zheng2022farl} deep feature space). We demonstrate through a series of both qualitative and quantitative experiments that our method is capable of anonymizing the identity of the images whilst--crucially--better-preserving the facial attributes. We make the code and the pre-trained models publicly available at: \url{https://github.com/chi0tzp/FALCO}.
\end{abstract}

\section{Introduction}\label{sec:intro}
    
    The ubiquitous use of mobile devices equipped with high-resolution cameras and the ability to effortlessly share personal photographs and videos on social media poses a significant threat to data privacy. Considering that modern machine learning algorithms learn from vast amounts of data often crawled from the Web~\cite{schuhmann2021laion400m,karras2019stylegan}, it has become increasingly important to consider the impact this has on the privacy of those individuals depicted. Motivated by privacy concerns, many societies have recently enacted strict legislation, such as the General Data Protection Regulation (GDPR)~\cite{GDPR}, which requires the consent of every person that might be depicted in an image dataset. Whilst such laws have obvious benefits to the privacy of those featured in image datasets, this is not without costly side effects to the research community. In particular, research fields such as computer vision and machine learning rely on the creation and sharing of high-quality datasets of images of humans for a number of important tasks including security~\cite{reid2018liu}, healthcare~\cite{foteinopoulou2021estimating, bishay2021schinet}, and creative applications~\cite{karras2019stylegan,rombach2021highresolution}.
    
    A recent line of research focuses on overcoming this issue by \textit{anonymizing} the identity of the individuals in image datasets. Through this approach, the machine learning community can still benefit from the wealth of large datasets of high-resolution images, but without cost to privacy. 
    
    \begin{figure}[t]
        \centering
        \includegraphics[width=0.47\textwidth]{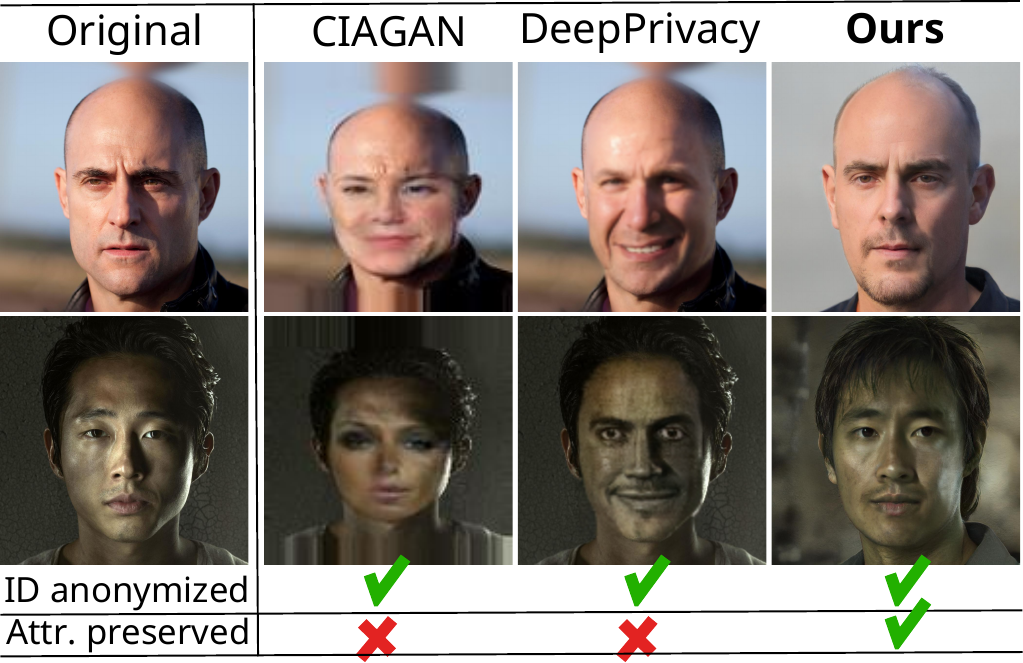}
        \caption{Comparison of the proposed method to CIAGAN~\cite{maximov2020ciagan} and DeepPrivacy~\cite{hukkelas2019deepprivacy} in terms of identity anonymization and attribute preservation.}
        \label{fig:summary}
    \end{figure}
    
    This research field has seen several developments throughout the last few years. Early methods proposed by the computer vision community attempt to solve this problem with simple solutions based on blurring~\cite{DU2019347} or other masking techniques, such as pixelation~\cite{GERSTNER2013333}. The result of this masking process succeeds in anonymizing the images by completely hiding the identity-related components, but as a consequence renders the facial attribute information such as a person's pose, expression, or skin tone (from which many computer vision tasks learn) indecipherable. Another problem with these methods is that, whilst the resulting images may not be re-identifiable by humans, they can often be reversed by deep learning models~\cite{Faceless_detection, defeating_obfuscation}.
    
    Another line of work leverages the power of Generative Adversarial Networks (GANs)~\cite{gan}, which have recently been used for discovering controllable generation paths in their latent or feature spaces~\cite{oldfield2022panda,tzelepis2022contraclip,oldfield2021tensor,tzelepis2021warpedganspace,bounareli2022stylemask}. Towards face anonymization, GANs have been incorporated in order to synthesize new images in order to obtain photos that maintain most of the image while changing the face of the subject of interest. In particular, these approaches use techniques like image inpainting~\cite{hukkelas2019deepprivacy}, conditional generation~\cite{maximov2020ciagan}, attribute manipulation~\cite{li2019anonymousnet}, or adversarial perturbation~\cite{shan2020fawkes}. These works are able to obtain anonymized images that can still be used for computer vision tasks such as tracking and detection, with very good results in terms of privacy preservation. However, many of these works lack the ability to generate natural-looking faces and often fail to preserve the original facial attributes in the anonymized images (or, on the occasions in which such methods do preserve the facial attributes, they fail to demonstrate this quantitatively). This is critical for many applications which rely on the attributes of the inner face, such as expression recognition~\cite{kollias2019deep}, or mental health affect analysis~\cite{niki2022affect}. To further complicate the picture, a fundamental problem often found with existing works is the way in which the anonymized images copy not just the original image's background, but also more identifiable features~\cite{maximov2020ciagan,hukkelas2019deepprivacy}, such as the clothes of an individual, or their hair  (see examples of this in Fig.~\ref{fig:summary}). We argue that leaving such structure of the images unchanged constitutes a glaring privacy vulnerability, as one can re-identify the original image from the anonymized counterpart by comparing the image background or person's clothes.
    
    Motivated by these concerns, in this work we propose to de-identify individuals in datasets of facial images whilst \textit{preserving} the facial attributes of the original images. To achieve this, in contrast to existing work~\cite{maximov2020ciagan, hukkelas2019deepprivacy, li2019anonymousnet, wu2018ppgan, wen2022identitydp} that train custom neural networks from scratch, we propose to work directly in the latent space of a powerful pre-trained GAN, optimizing the latent codes directly with losses that explicitly aim to retain the attributes and obfuscate the identities. More concretely, we use a deep feature-matching loss~\cite{zheng2022farl} to match the high-level semantic features between the original and the fake image generated by the latent code, and a margin-based identity loss to control the similarity between the original and the fake image in the ArcFace~\cite{Deng2021arcface} space. The initialisation of the latent codes is obtained by randomly sampling the latent space of GAN, using them to generate the corresponding synthetic images and finding the nearest neighbors in a semantic space (FARL~\cite{zheng2022farl}). In order to preserve texture and pose information of the original image, we perform inversion of the original image and retain the parts that correspond to the properties we want to preserve in the final code. This results in a latent code that yields a high-resolution image that contains a new identity but retains the same facial attributes as the original image.
    
    The main contributions of this paper can be summarized as follows:
    \begin{itemize}
        \item To the best of our knowledge, we are the first to address the problem of identity anonymization whilst also explicitly retaining facial attributes.
        \item We propose a novel methodology and loss functions working with \textit{pre-trained} GANs capable of generating high-resolution anonymized datasets.
        \item We show through a series of thorough experiments on both Celeba-HQ~\cite{liu2015faceattributes} and LFW~\cite{LFWTech} that our method competes with the state-of-the-art in obfuscating the identity, whilst better-retaining the facial attributes under popular quantitative metrics.
    \end{itemize}

\section{Related Work}\label{sec:related_work}
    
    \textbf{Face obfuscation} The first privacy-preserving approaches proposed were based on obfuscating the face of the person. This means that different techniques, like blurring, masking, or pixelating~\cite{boyle2000filter,Chen2007ToolsFP,neustaedter2006,tansuriyavong2001masking} are used to completely remove the personally identifiable information (PII). In the masking approach, the face region is simply covered with a shape such that the body or face of the person is completely covered, with pixelation the resolution of the face region is reduced, and blurring uses Gaussian filters with varying standard deviation values, allowing different strengths of the blurring. Tansuriyavong et al.~\cite{tansuriyavong2001masking} de-identifies people in a room by detecting the silhouette of the person, masking it, and showing only the name to balance privacy protection and the ability to convey information about the situation, Chen et al.~\cite{Chen2007ToolsFP} obscures the body information of a person with an obscuring algorithm exploiting the background subtraction technique leaving only the body outline visible. Naive de-identification techniques that maintain little information about the region of interest, such as pixelation and blurring, may seem to work to the human eye, but there exist approaches able to revert the anonymized face to its original state~\cite{Faceless_detection, defeating_obfuscation}. To improve the level of privacy protection, techniques defined as $k$-Same have been introduced~\cite{newton2005preserving}, where, given a face, a de-identified visage is computed as the average of the $k$ closest faces and then used to replace the original faces from the ones used in the calculation. This set of  techniques works very well in removing privacy-related information, however, the result of the process completely removes the information related to the facial region, resulting in samples that are impossible to use in applications that need to use face detectors, trackers, or facial attributes. To solve these issues, our method instead leverages the generation capability of the state-of-the-art StyleGAN2~\cite{stylegan2_karras20cvpr} to obtain realistic-looking face images, which are still detectable and that retain the facial attributes present in the original image.
    
    \begin{figure*}[t]
        \centering
        \includegraphics[width=0.99\textwidth]{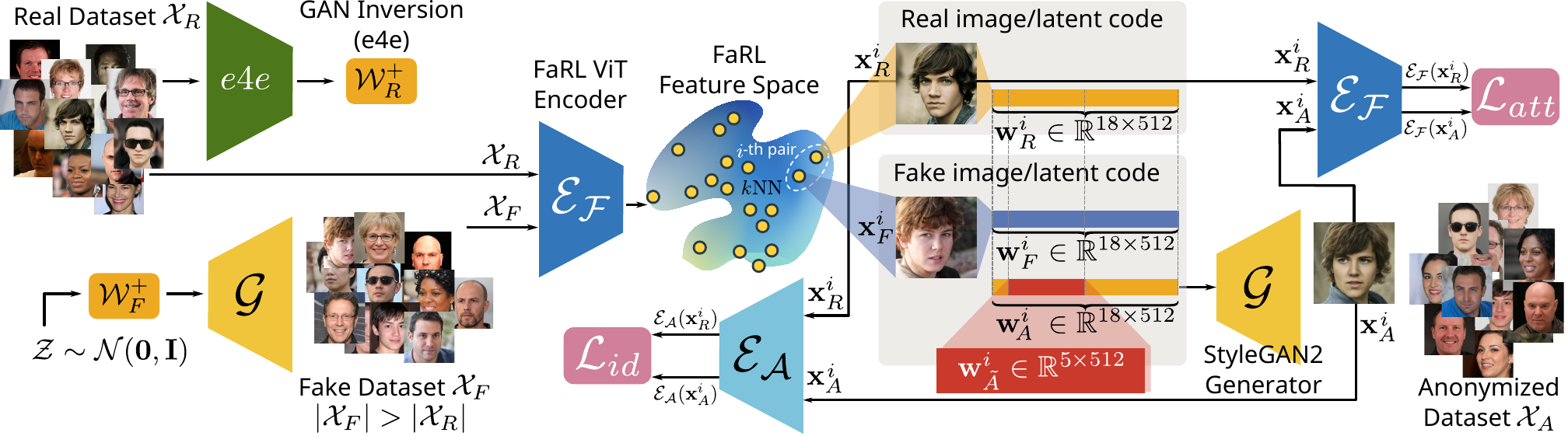}
        \caption{
        Overview of the proposed method: optimizing the trainable portion of the latent code $\mathbf{w}^i_{\Tilde{A}}\in\mathbb{R}^{5\times 512}$ to obfuscate the identity of the resulting synthetic image $\mathbf{x}^i_A$ with $\mathcal{L}_{id}$ whilst preserving the facial attributes with $\mathcal{L}_{att}$.
        }
        \label{fig:overview}
    \end{figure*}
    
    \textbf{Generative face anonymization} After the advent of GANs~\cite{gan}, several lines of work have been proposed to tackle the problem of anonymization leveraging the generative power of these networks. Prior to this,~\cite{cho2020cleanir,ma2021cfanet} proposed auto-encoder based methods, in particular Cho et al.~\cite{cho2020cleanir} used such networks to learn to disentangle the identity information from the attributes given a vector representation of an image, allowing then to tweak the identity part of the vector to obtain an anonymized subject. This work reported emotion preservation results, however in this case we are concerned with the preservation of facial attributes more generally. The main weakness of such methods is the lack of sharpness of the generated images. Given the power of StyleGAN2 to synthesize sharp, high-resolution images, our framework avoids the problem of generating blurry images. Several methods~\cite{maximov2020ciagan,wen2022identitydp,shan2020fawkes,hukkelas2019deepprivacy,li2019anonymousnet,wu2018ppgan,dallasen2021anony} also employ the use of GAN networks to similar ends, given their incredible ability to capture the distribution of the training samples and then generate similar looking images. DeepPrivacy~\cite{hukkelas2019deepprivacy} extracts the face region along with sparse facial key points from a face image, removes the face from the image using the bounding box coordinates and sets its region to a constant value. This is passed to a conditional GAN~\cite{mirza2014cgan}, along with background and pose information, which inpaints a randomly generated face from StyleGAN2's~\cite{stylegan2_karras20cvpr} generator, while maintaining contextual and pose information. CIAGAN~\cite{maximov2020ciagan} also uses a conditional GAN, performing a form of conditional ID-swapping. The model uses an identity discriminator to force the generated image, conditioned on the landmark information and masked image, to display a different identity from the source one with similar features to the borrowed identity. These two methods obtain great privacy preservation results, but still lack the ability to generate natural-looking faces for images of large resolution. Even if contextual knowledge is injected under the form of the masked original image, i.e., without the face region, there is no guarantee that facial attributes are retained after the anonymization. In the case of CIAGAN in particular, the results are visually similar to the original subject only when the conditional ID happens to share the same features such as gender, age or skin tone. Using our attribute preservation loss, and thanks to ``prior knowledge'' of textural information that we gain from the original inverted image latent code, these issues are solved. Techniques of ID-swapping, like ~\cite{maximov2020ciagan}, present also another issue: since \textit{real} images' identity features are used to condition the anonymized face, it is unclear if this truly solves the privacy problem. These issues are avoided in our framework, as the pre-trained generator outputs random, non-existing faces and thus no information about the original identities is retained. One of the most recent works, namely IdentityDP~\cite{wen2022identitydp}, tackles the problem of anonymization in a three-fold process: first an attribute encoder and an identity encoder are used to extract the corresponding features, which are then injected into a GAN to reconstruct the original image. In this way the encoders learn to disentangle attribute and identity information. Then, the identity vector is perturbed with Laplacian noise and, finally, it is passed to the generator, along with the attributes vector to obtain the de-identified image. In the absence of any publicly available implementation of~\cite{wen2022identitydp}, we do not provide quantitative comparisons with~\cite{wen2022identitydp}. We can, however, comment on their qualitative results, stating that, indeed, the facial attributes are maintained, but the resulting image can still be recognized by a human observer. Our proposed method does not suffer from such a problem, since the controllable similarity allows us to obtain face images of completely different persons that share the same attributes as the original subject.

\section{Proposed Method}\label{sec:proposed_method}
    
    In this section, we present our method for anonymizing the identity of faces in a given real face dataset by optimizing the representations of the dataset's images in the latent space of a pre-trained StyleGAN2~\cite{stylegan2_karras20cvpr}. More specifically, given a real dataset $\mathcal{X}_R$, we first create a fake dataset $\mathcal{X}_F$ by randomly generating a large set (i.e., such that $\vert\mathcal{X}_F\vert>\vert\mathcal{X}_R\vert$) of fake images and obtaining the corresponding latent codes in the $\mathcal{W}^+$ space of StyleGAN2, namely, $\mathcal{W}^+_F$. Additionally, we obtain the latent codes of the real dataset in the $\mathcal{W}^+$ space by inverting its images using e4e~\cite{2021e4e}, arriving at a set of latent codes $\mathcal{W}^+_R$. In order to obtain meaningful initial values for the latent codes that will be optimized to create the anonymized version of the real dataset, namely $\mathcal{X}_A$, we first pair the real images from the original set (i.e., $\mathcal{X}_R$) with fake ones from the generated dataset (i.e., $\mathcal{X}_F$) in the feature space of the ViT-based FaRL~\cite{zheng2022farl} image encoder and use their latent codes for initializing the aforementioned trainable codes. The latent codes of the anonymized dataset are then optimized under the following objectives via two novel loss functions: (a) to be similar to the corresponding real ones, up to a certain margin, using the proposed identity loss ($\mathcal{L}_{id}$), and (b) to preserve the facial attributes of the corresponding real ones by being pulled closer in the feature space of the pre-trained FaRL~\cite{zheng2022farl} image encoder using the proposed attribute preservation loss ($\mathcal{L}_{att}$). In this way, in contrast to state-of-the-art works~\cite{maximov2020ciagan,hukkelas2019deepprivacy}, the anonymized images are optimized to inherit the labels of the original ones. An overview of the proposed method is given in Fig.~\ref{fig:overview}.
        
    The rest of this section is organized as follows: in Sect.~\ref{subsec:background} we briefly introduce the pre-trained modules of our framework, in Sect.~\ref{subsec:fake_dataset_pairing} we discuss the initialization of the latent codes that will generate the anonymized version of the real dataset, and in Sect.~\ref{subsec:latent_code_optim} we present the proposed optimization process and losses.
    
    \subsection{Background}\label{subsec:background}
        
        \textbf{StyleGAN2} We use the StyleGAN2's~\cite{stylegan2_karras20cvpr} generator $\mathcal{G}$, pre-trained on the FFHQ~\cite{stylegan2_karras20cvpr} dataset and, in particular, its $\mathcal{W}^+$ latent space. In this case, we operate on the latent codes $\mathbf{w}\in\mathbb{R}^{18\times 512}$, where the first 9 layers are responsible for coarse- and medium-grained attributes (such as the head pose and facial texture details), while the rest correspond to more fine-grained attributes (such as the hair colour, or the skin tone, as first identified in~\cite{karras2019stylegan}).
        
        \textbf{e4e} For the inversion of real images onto the $\mathcal{W}^+$ space of StyleGAN2, we use e4e~\cite{2021e4e}, which has been trained in order to preserve a good trade-off between fidelity of the inversion and editability in $\mathcal{W}^+$.
        
        \textbf{ArcFace} For measuring the similarity of the identities of two face images, we use ArcFace~\cite{Deng2021arcface}, which represents images in a $512$-dimensional identity-related feature space using which we optimize the GAN latent codes to generate images to maximize the cosine similarity between features corresponding to the same face identity.
        
        \textbf{FaRL} For the representation of images in a meaningful and rich semantic feature space, we use FaRL~\cite{zheng2022farl}, a universal facial representation scheme trained in a contrastive manner in 20 million face images-text pairs. Specifically, we use the ViT image encoder of the FaRL framework in order to represent images in a $512$-dimensional feature space and find meaningful initial values for the latent codes that will be optimized to anonymize the real dataset.
    
    \subsection{Fake dataset generation and pairing with real images}\label{subsec:fake_dataset_pairing}
        
        Given a dataset $\mathcal{X}_R$ of real face images, we incorporate the generator $\mathcal{G}$ of a StyleGAN2~\cite{stylegan2_karras20cvpr} pre-trained on the FFHQ~\cite{stylegan2_karras20cvpr} dataset, in order to generate a set of fake face images $\mathcal{X}_F$, where $\vert\mathcal{X}_F\vert>\vert\mathcal{X}_R\vert$. We do so by sampling from the $\mathcal{Z}$ latent space of StyleGAN2, i.e., the Standard Gaussian $\mathcal{N}(\mathbf{0},\mathbf{I})$, and by then obtaining the corresponding $\mathcal{W}^+$ latent codes (using the input MLP of $\mathcal{G}$), i.e., the set $\mathcal{W}^+_F$. At the same time, we calculate the latent representations of the face images in the real dataset $\mathcal{X}_R$ by inverting them using e4e~\cite{2021e4e}. This assigns $\mathcal{X}_R$ with the set of corresponding $\mathcal{W}^+$ latent codes, i.e., the set $\mathcal{W}^+_R$. This is illustrated in the left part of Fig.~\ref{fig:overview}.
        
        For pairing the real images in $\mathcal{X}_R$ with fake ones in $\mathcal{X}_F$ we use the pre-trained FaRL~\cite{zheng2022farl} ViT-based image encoder $\mathcal{E}_{\mathcal{F}}$ and we represent all images of each dataset using the class (i.e., CLS) token representation, i.e., in a $512$-dimensional features space. By doing so, we obtain a powerful feature representation of both datasets, which we subsequently use in order to train a $k$NN classifier and obtain, for each real image, the closest fake one in terms of the Euclidean distance. More formally, after the aforementioned process of generation and pairing, the images/latent codes of the real dataset $\mathcal{X}_R$ are paired with images/latent codes in the fake dataset $\mathcal{X}_F$ forming the following set of pairs
        \begin{align}\label{eq:pairs}
            \{\left((\mathbf{x}^i_R,\mathbf{w}^i_R),(\mathbf{x}^i_F,\mathbf{w}^i_F)\right)\colon & \mathbf{x}^i_R\in\mathcal{X}_R,\mathbf{w}^i_R\in\mathcal{W}^+_R,\\ \nonumber
            & \mathbf{x}^i_F\in\mathcal{X}_F,\mathbf{w}^i_F\in\mathcal{W}^+_F,\\ \nonumber
            & i=1,\ldots,\vert\mathcal{X}_R\vert\}.
        \end{align}
        
        In order to initialize the latent codes that will be optimized to anonymize the real images, we use the above pairs of real-fake latent codes as follows. Given the $i$-th real image, we first modify the latent code of the corresponding fake one (i.e., its nearest neighbor), $\mathbf{w}^i_F\in\mathbb{R}^{18\times512}$, and replace layers 3-7 with a trainable vector $\mathbf{w}^i_{\Tilde{A}}\in\mathbb{R}^{5\times512}$, while we set its first three layers (i.e., layers 0-2) and the last layers (i.e., layers 8-17) equal to the corresponding layers of the real latent code $\mathbf{w}^i_R$. By doing so, we arrive at a latent code $\mathbf{w}^i_A\in\mathbb{R}^{18\times512}$, initialized so as a) we retain information that is crucial for generating anonymized face images with head pose and other coarse geometric details same as the corresponding real ones (layers 0-2), b) we maintain the color distribution and background information (layers 8-17) of the real ones, and c) optimize information that is critical for the identity characteristics of a face (layers 3-7). This is illustrated in the centre part of Fig.~\ref{fig:overview}.

    \subsection{Latent code optimization}\label{subsec:latent_code_optim}
        
        In order to create an anonymized version $\mathcal{X}_A$ of the real dataset $\mathcal{X}_R$, we use the pairs of real-fake images obtained and initialized by the process discussed in the previous section and shown in (\ref{eq:pairs}), i.e., pairs of real and fake images that are semantically close to each other in terms of the FaRL image representation scheme. More specifically, the real image of each pair, $\mathbf{x}^i_R$, along with the corresponding anonymized image, $\mathbf{x}^i_A$, generated by the modified latent code, $\mathbf{w}^i_A$, are used for calculating the proposed losses. That is, the identity loss $\mathcal{L}_{id}(\mathbf{x}^i_A,\mathbf{x}^i_R)$ so as $\mathbf{x}^i_A$ retains a similar identity to $\mathbf{x}^i_R$, up to a desired margin, and the attribute preservation loss $\mathcal{L}_{att}(\mathbf{x}^i_A,\mathbf{x}^i_R)$ that imposes that the facial attributes of the original image are preserved in the anonymized image.
        
        Given a pair consisting of a real image $\mathbf{x}^i_R$ and its anonymized version $\mathbf{x}^i_A$, we estimate the learnable parts of its latent code $\mathbf{w}^i_{\Tilde{A}}\in\mathbb{R}^{8\times512}$, for  $i=1,\ldots,\vert\mathcal{X}_R\vert$ by optimizing the following losses:
    
        \paragraph{Identity loss} The identity loss is defined as follows
        \begin{equation}\label{eq:id_loss}
            \mathcal{L}_{id}(\mathbf{x}^i_A,\mathbf{x}^i_R)
            =
            \left\vert
            \cos\left(\mathcal{E}_{\mathcal{A}}(\mathbf{x}^i_A),\mathcal{E}_{\mathcal{A}}(\mathbf{x}^i_R)\right)
            -m\right\vert,
        \end{equation}
        where $\cos(\cdot,\cdot)$ denotes the cosine distance, $\mathcal{E}_{\mathcal{A}}$ denotes the ArcFace~\cite{Deng2021arcface} identity encoder, and $m$ denotes a hyperparameter that controls the dissimilarity between the real and the anonymized face images. When $m=0$, the proposed identity loss imposes orthogonality between the features of the real and the anonymized face images, leading to anonymized faces with large identity difference compared to the corresponding real ones. By contrast, when $m=1$, the proposed identity loss imposes high similarity between the features of the real and the anonymized face images. That is, the hyperparameter $m$ controls the trade-off between data utility and privacy preservation.
        
        \paragraph{Attribute preservation loss} The attribute preservation loss is defined as follows
        \begin{equation}\label{eq:att_loss}
            \mathcal{L}_{att}(\mathbf{x}^i_A,\mathbf{x}^i_R)
            =
            \left\lVert
            \mathcal{E}_{\mathcal{F}}(\mathbf{x}^i_A)-\mathcal{E}_{\mathcal{F}}(\mathbf{x}^i_R)
            \right\rVert_1,
        \end{equation}
        where $\mathcal{E}_{\mathcal{F}}$ denotes the FaRL~\cite{zheng2022farl} ViT-based image encoder. It is worth noting that we found empirically that using the patch-level features of the ViT (i.e., the $14\times14$ $512$-dimensional features, flattened as $14\cdot14\cdot512$-dimensional vectors, leads to better attribute preservation than using the features at the class (CLS) token. We argue that maintaining the raw representation allows for better results compared to using only the class token, as this encodes a class contextual representation of the image, while the untouched patches' features contain a higher degree of information.

\section{Experiments}\label{sec:experiments}

    In this section we evaluate the performance of our anonymization framework against other state-of-the-art anonymization works, evaluating our results on privacy-related metrics in Sect.~\ref{sec:exp:privacy}, and -- in contrast to other works -- attribute classification metrics in Sect.~\ref{subsubsec:attribute}. Finally, we show in Sect.~\ref{sec:ablation} the impact of the identity loss margin involved in our method through an ablation study.
    
    \paragraph{Datasets} We perform anonymization on the following datasets: (i) \textbf{CelebA-HQ}~\cite{liu2015faceattributes}, which contains 30000 $1024\times1024$ face images of celebrities from the CelebA dataset with various demographic attributes (e.g., age, gender, race) and where each image is annotated with 40 attribute labels related to the inner and outer regions of the face, and (ii) \textbf{LFW}~\cite{LFWTech}, which contains over 13000 images collected from the Web (5749 identities with 1680 of those identities being pictured in at least 2 images).
    
    \paragraph{State of the art} We compare our anonymization framework with two state-of-the-art anonymization methods, namely CIAGAN~\cite{maximov2020ciagan} and DeepPrivacy~\cite{hukkelas2019deepprivacy}.

    \subsection{Evaluation metrics}\label{sec:evaluation-metrics}
        We evaluate our method by quantifying privacy preservation, image quality, and attribute preservation. We briefly introduce the metrics we use below:

        \paragraph{Image quality and identity anonymization} We quantify the ability to anonymize images by measuring the ``re-identification rate'' (defined as the number of images whose identity is still detected in the anonymized version, over the total number of images) using FaceNet~\cite{schroff2015facenet}, pre-trained on two large-scale face datasets (CASIA WebFace~\cite{yi2014casia} and VGGFace2~\cite{cao2017vggface}). Moreover, we measure the ``detection rate'' as the number of anonymized images for which a valid face is successfully detected over the total number of images in the dataset. By quantifying how recognisable a face is to a machine learning algorithm, this metric is an important measure of the quality of the facial image. To measure this, we use the MTCNN~\cite{zhang2016mtcnn} face detector. An ideal anonymization method would retain a valid face in all anonymized images (100\% detection rate), but anonymize all the particular identities (0\% re-identification rate). Finally, we report the Fr\'echet Inception Distance (FID)~\cite{heusel2017fid} for all generated images as a measure of the quality of the anonymized datasets.
        
        \paragraph{Attribute preservation} Unlike other works~\cite{maximov2020ciagan,hukkelas2019deepprivacy}, we propose a protocol to quantify how well each method can retain the attributes of the original images. More specifically, the evaluation is posed as a standard classification task and the metric used to quantify this is the accuracy of classifiers on the real test sets when trained on the anonymized training set. In this way, one can quantify how well the anonymized training data has retained the original attributes in the images. The train/test split structure followed is the one provided by the official CelebA dataset in the case of CelebA-HQ~\cite{liu2015faceattributes}, while the images from LFW~\cite{LFWTech} are randomly shuffled and then split with an 80-20 ratio. We use a MobileNetV2~\cite{sandler2018mobilenetv2} to perform multi-label classification, trained with a focal loss~\cite{2017focal} to handle class imbalance.

    \subsection{Comparison to state-of-the-art (SOTA)}\label{sec:soa_comparisons}
        In this section, we report the evaluation performance of our method compared to two other SOTA methods (CIAGAN~\cite{maximov2020ciagan} and DeepPrivacy~\cite{hukkelas2019deepprivacy}) using the evaluation metrics introduced earlier. Finally, in Sect.~\ref{sec:qualitative} we conduct a qualitative comparison to the SOTA.

        \subsubsection{Image quality and De-identification}\label{sec:exp:privacy}
            
            In Tables~\ref{tab:chq_privacy_results}, \ref{tab:lfw_privacy} we show the results for FID, face detection, and face re-identification for the two considered datasets. We see that our method excels at producing the most realistic-looking images under the FID metric for CelebA-HQ in Tab.~\ref{tab:chq_privacy_results}, and also outperforms the baselines for the FID metric on LFW~\cite{LFWTech} in Tab.~\ref{tab:lfw_privacy} when considering the CelebA-HQ~\cite{liu2015faceattributes} dataset as the ``target'' distribution\footnote{Given that CelebA-HQ is of much higher quality than LFW, we report both cases to demonstrate that our images can better match the distribution of high-resolution data.}. We argue this success is due to the way in which we design our method to operate in the latent space of a well-trained GAN, capable of producing high-resolution, sharp images. On the other hand, the existing SOTA involves techniques such as image inpainting, which we find have a tendency to introduce small artifacts in the anonymization procedure. The realism of our generated images is further attested to by the perfect face detection scores in both Tables~\ref{tab:chq_privacy_results} and \ref{tab:lfw_privacy} -- indicating the images contain recognisable faces readily usable for downstream machine learning tasks.
    
            However, our images are not just of high quality, but also successfully anonymize the identity--we also see from the last columns of Tables~\ref{tab:chq_privacy_results} and \ref{tab:lfw_privacy} that our anonymization results are competitive with the SOTA. However, it is important to note that whilst the baselines excel under this metric, they fail to preserve the attributes to the extent of our method, which we detail in the next section.
    
            \begin{table}[t]
            \resizebox{\columnwidth}{!}{ 
            \begin{tabular}{l|ccccccc}
             &
              FID$\downarrow$ &
              \multicolumn{2}{c}{Detection$\uparrow$} &
              \multicolumn{2}{c}{Face re-ID$\downarrow$}\\ 
                              &         & dlib       & MTCNN(\%)          & CASIA(\%)           & VGG(\%)\\             
            \hline
            Randomly generated      &\underline{18.09}    & 100   &99.99\       &3.61       &1.08\\
            CIAGAN~\cite{maximov2020ciagan}            & 37.94      & 95.10    & 99.82        & \textbf{2.19} & \textbf{0.37}\\
            DeepPrivacy~\cite{hukkelas2019deepprivacy}       & 32.99    & 92.82      & 99.85        & 3.61          & 1.05\\
            \textbf{Ours (ID)} & 44.12   & 98.58       & 97.99        & 3.28                & 0.58\\
            \textbf{Ours (ID+attributes)} &
              44.11 &       100
              & \textbf{100} &      3.06 &     2.06 &\\
            \textbf{Ours}      & \textbf{29.93}  & 100 & \textbf{100} & 2.80          & 1.67\\
            \end{tabular}}%
            \caption{CelebA-HQ~\cite{liu2015faceattributes} privacy and image quality results.}
            \label{tab:chq_privacy_results}
            \end{table}
            
            \begin{table}[t]
            \resizebox{\columnwidth}{!}{%
            \begin{tabular}{l|ccccccc}
             & FID$\downarrow$   &FID (C-HQ)$\downarrow$ &\multicolumn{2}{c}{Detection$\uparrow$}       & \multicolumn{2}{c}{Face re-ID$\downarrow$}\\
             &&     & dlib  &MTCNN(\%) & CASIA(\%)    & VGG(\%)\\
            \hline
            CIAGAN~\cite{maximov2020ciagan} 
            &\textbf{22.07} 
            & 85.23 & 98.14 & 99.89  &\textbf{0.17}  &\textbf{0.91}\\
            DeepPrivacy~\cite{hukkelas2019deepprivacy} 
            &23.46 
            & 123.67 & 96.70  &99.57  &2.74  &1.52\\
            \textbf{Ours} 
            &27.45 
            &\textbf{68.88} & 100 &\textbf{100}  &2.07  &1.58\\
            \end{tabular}}%
            \caption{LFW~\cite{LFWTech} privacy and image quality results.}
            \label{tab:lfw_privacy}
            \end{table}

        \subsubsection{Attribute preservation}\label{subsubsec:attribute}
            In this section we quantify the attribute preservation of the anonymization methods:
    
            \paragraph{CelebA-HQ} For CelebA-HQ~\cite{liu2015faceattributes}, which provides images annotated according to 40 facial attributes, we first train a MobileNetV2~\cite{sandler2018mobilenetv2} on the anonymized training sets to predict the attributes of the images, and evaluate its performance on the untouched test set--as a proxy measure for how well the anonymized images have retained the original expected facial attribute labels. Tab.~\ref{tab:accuracy_results} shows the performances of our framework compared to the other methods and also when training using the original dataset.

            As can be seen, our method's images result in a classifier capable of almost the same accuracy as when training on the original labels, demonstrating the ability of our method to retain the original facial features. Whilst the other two baselines also produce reasonable results under this \textit{combined} accuracy metric, we argue this is because of the way in which they preserve the image outside the region of the inner face of the images -- out of 40 attributes, 17 correspond to the ``outer face'' region. As shown in Tab.~\ref{tab:accuracy_inner_outerface} with the accuracy breakdown for the individual attributes, the face inpainting methods excel at preserving the "outer face" attributes as expected, whereas we often outperform the baselines for attributes related to the ``inner'' region of the face, such as ``eyeglasses'' or ``smile''.

            \paragraph{LFW} Since no official annotations regarding facial attributes are provided for the LFW~\cite{LFWTech} dataset, we instead use two classifiers~\cite{2021anycost,2021talk2edit} pretrained on CelebA-HQ~\cite{liu2015faceattributes}. The model of~\cite{2021anycost} predicts all the 40 attributes officially provided by CelebA, while~\cite{2021talk2edit} predicts only 5 of them, namely \textit{Bangs, Eyeglasses, No\_Beard, Smiling} and \textit{Young}  (more details on these two classifiers can be found in the supplementary material~\ref{supp:attr_classifiers}). For the original LFW~\cite{LFWTech} dataset, we first predict ``pseudo-labels'' to approximate the ground-truth attribute labels using~\cite{2021anycost,2021talk2edit}, and then proceed with the same classification procedure as above. As we can see from the first column of Tab.~\ref{tab:pseudo_accuracy}, the accuracy results when training on pseudo-labels on CelebA-HQ~\cite{liu2015faceattributes} are close to those using the real labels in Tab.~\ref{tab:accuracy_results}, validating the reliability of the classifiers to generate accurate pseudo-labels. Furthermore, we see in the last two columns of Tab.~\ref{tab:pseudo_accuracy} that our method is able to generate images that much better preserve the facial attributes of the original images than the existing SOTA anonymization methods, through being able to train more accurate attribute classifiers.
            
            \begin{table}[t]
            \resizebox{\columnwidth}{!}{%
            \begin{tabular}{l|ccc}
                &Inner face &Outer face &Combined\\ 
            \hline
            Original     &\underline{0.8409}    &\underline{0.8683}   &\underline{0.8539} \\ 
            CIAGAN~\cite{maximov2020ciagan}       &0.7277    &0.8372   &0.7852 \\
            DeepPrivacy~\cite{hukkelas2019deepprivacy}  &0.7658    &0.8511   &0.8135 \\
            \textbf{Ours} &\textbf{0.7817}    &\textbf{0.8518}   &\textbf{0.8181} \\
            \end{tabular}%
            }
            \caption{Attribute classification results on CelebA-HQ~\cite{liu2015faceattributes}.}
            \label{tab:accuracy_results}
            \end{table}
            
            \begin{table}[t]
            \resizebox{\columnwidth}{!}{%
            \begin{tabular}{l|ccccc}
                                 & FID   & Detection & \multicolumn{2}{c}{Face re-ID} & Accuracy \\
                                 &       & MTCNN(\%) & CASIA(\%)       & VGG(\%)      &              \\
            \hline
            \textbf{Ours (m=.0)} & 29.93 & \textbf{100}       & \textbf{2.80}            & \textbf{1.67}         & 0.8181       \\
            \textbf{Ours (m=.9)} & \textbf{27.58}    & \textbf{100}         & 3.41                 & 1.76             & \textbf{0.83}            
            \end{tabular}%
            }
            \caption{Ablation study on the margin $m$ on CelebA-HQ~\cite{liu2015faceattributes}.}
            \label{tab:ablation-2}
            \end{table}

            \begin{table}[t]
            \resizebox{\columnwidth}{!}{%
            \begin{tabular}{l|ccc}
            \textbf{}             & \textbf{CIAGAN}~\cite{maximov2020ciagan} & \textbf{DeepPrivacy}~\cite{hukkelas2019deepprivacy} & \textbf{Ours}   \\ \hline
            \textbf{Outer face region} &&&\\
            Bald               & \textbf{0.9778} & 0.9772          & 0.9769          \\
            Bangs              & 0.8127          & \textbf{0.85}   & 0.8241          \\
            Black\_Hair        & \textbf{0.7927} & 0.7794          & 0.7864          \\
            Blond\_Hair        & 0.8497          & \textbf{0.8708} & 0.8707          \\
            Brown\_Hair        & \textbf{0.7626} & 0.7615          & 0.7593          \\
            Double\_Chin       & \textbf{0.9377} & 0.9362          & 0.9364          \\
            Gray\_Hair         & \textbf{0.9603} & 0.9569          & 0.9587          \\
            Receding\_Hairline & \textbf{0.9168} & 0.9126          & 0.9117          \\
            Sideburns          & 0.9197          & \textbf{0.9228} & 0.9186          \\
            Straight\_Hair     & 0.53            & \textbf{0.7738} & 0.7702          \\
            Wavy\_Hair         & 0.6433          & 0.6603          & \textbf{0.6652} \\
            Wearing\_Earrings  & 0.6972          & 0.6721          & \textbf{0.7123} \\
            Wearing\_Hat       & 0.9636          & \textbf{0.9641} & 0.9595          \\
            Wearing\_Necklace  & \textbf{0.822}  & 0.8017          & 0.81            \\
            Wearing\_Necktie   & \textbf{0.9288} & 0.9281          & 0.9273          \\
            Oval\_Face         & \textbf{0.7938} & 0.7796          & 0.7783          \\
            Chubby             & \textbf{0.9247} & 0.922           & 0.9153          \\ \hline
            \textbf{Inner face region} &&&\\
            5\_o\_Clock\_Shadow   & 0.8579          & 0.8604               & \textbf{0.8711} \\
            Arched\_Eyebrows      & 0.6057          & 0.658                & \textbf{0.6684} \\
            Bags\_Under\_Eyes     & 0.6946          & 0.7156               & \textbf{0.7158} \\
            Big\_Lips             & 0.6167          & 0.5901               & \textbf{0.6194} \\
            Big\_Nose             & 0.6814          & \textbf{0.7228}      & 0.7182          \\
            Bushy\_Eyebrows       & 0.774           & \textbf{0.8288}      & 0.8267          \\
            Eyeglasses            & 0.9483          & \textbf{0.9622}      & 0.9564          \\
            Goatee                & 0.9284          & 0.9289               & \textbf{0.9303} \\
            Heavy\_Makeup         & 0.6197          & \textbf{0.7492}      & 0.6859          \\
            High\_Cheekbones      & 0.5356          & 0.668                & \textbf{0.6729} \\
            Male                  & 0.6891          & 0.7917               & \textbf{0.8381} \\
            Mouth\_Slightly\_Open & 0.5722          & 0.603                & \textbf{0.6305} \\
            Mustache              & 0.9398          & \textbf{0.9401}      & 0.9323          \\
            Narrow\_Eyes          & \textbf{0.8925} & 0.8839               & 0.883           \\
            No\_Beard             & 0.5359          & 0.5571               & \textbf{0.7615} \\
            Pale\_Skin            & 0.9418          & 0.9446               & \textbf{0.9451} \\
            Pointy\_Nose          & 0.6239          & 0.6291               & \textbf{0.6689} \\
            Rosy\_Cheeks          & \textbf{0.8826} & 0.8553               & 0.8825          \\
            Smiling               & 0.5607          & 0.6505               & \textbf{0.6666} \\
            Wearing\_Lipstick     & 0.6234          & \textbf{0.7721}      & 0.7579          \\
            Young                 & 0.7583          & 0.7706               & \textbf{0.7848} \\ \hline
            \end{tabular}%
            }
            \caption{Accuracy of attributes (inner and outer face regions).}
            \label{tab:accuracy_inner_outerface}
            \end{table}
            
            \begin{table*}[t]
            \resizebox{\linewidth}{!}{%
            \begin{tabular}{l|ccc}
                                 & CelebA-HQ (labels from~\cite{2021anycost}) & LFW (labels from~\cite{2021anycost}) & LFW (labels from~\cite{2021talk2edit}) \\
            \hline
            \textbf{CIAGAN~\cite{maximov2020ciagan}}            & 0.7721            & 0.9143            & 0.7045       \\
            \textbf{DeepPrivacy~\cite{hukkelas2019deepprivacy}} & 0.7902            & 0.9133            & 0.7019      \\
            \textbf{Ours}                                        & \textbf{0.8215}   & \textbf{0.9157}   & \textbf{0.7209}       
            \end{tabular}%
            }
            \caption{Accuracy on CelebA-HQ~\cite{liu2015faceattributes} and LFW~\cite{LFWTech} of anonymized faces using the pseudo-labels generated by the classifiers of~\cite{2021anycost,2021talk2edit}.}
            \label{tab:pseudo_accuracy}
            \end{table*}
            
            \begin{figure}[t]
                \centering
                \includegraphics[width=\linewidth]{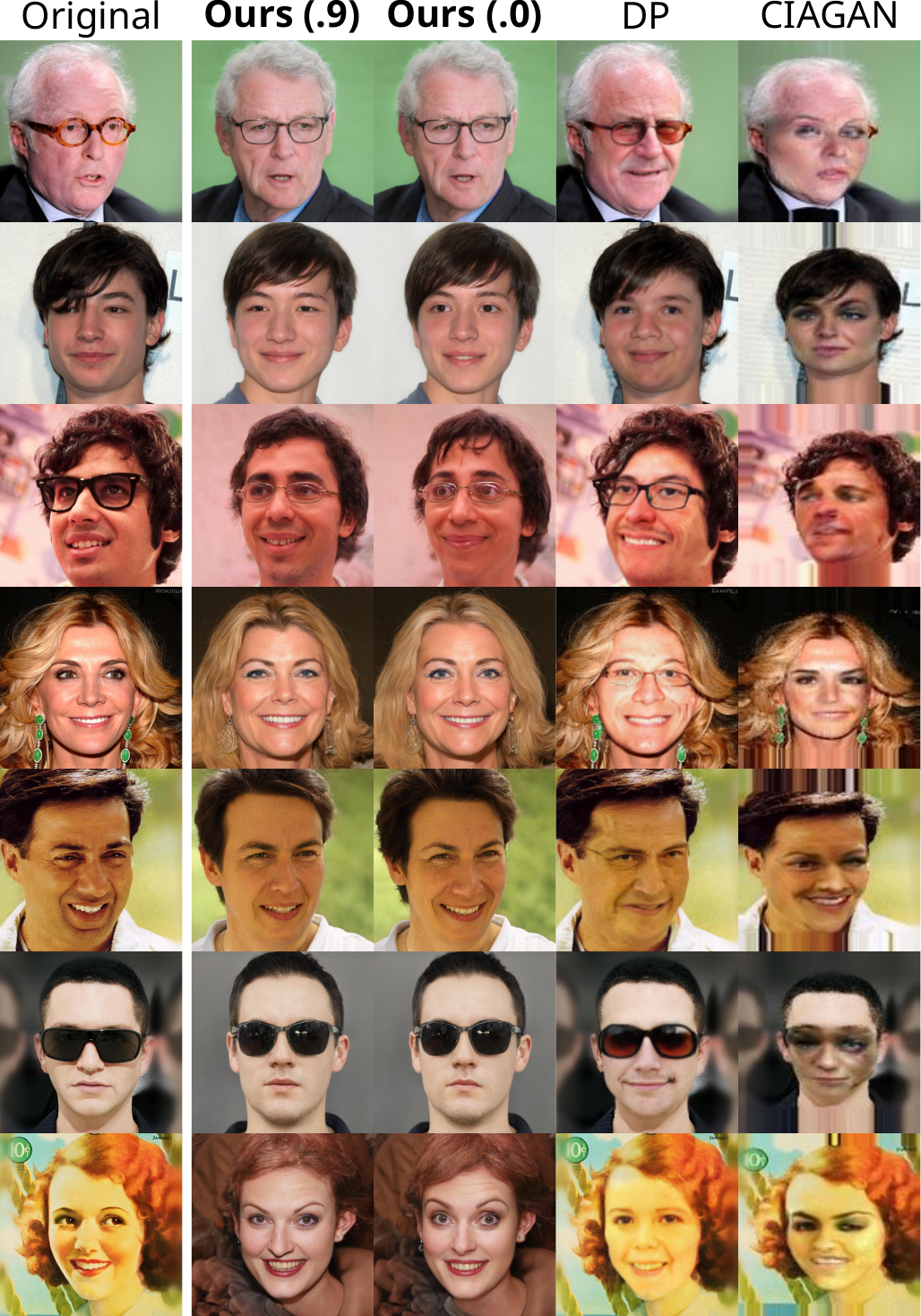}
                \caption{Anonymization results on CelebA-HQ~\cite{liu2015faceattributes} in comparison to DeepPrivacy (DP)~\cite{hukkelas2019deepprivacy} and CIAGAN~\cite{maximov2020ciagan}.}
                \label{fig:celebahq_qual}
            \end{figure}
            
            \begin{figure}[t]
                \centering
                \includegraphics[width=0.95\linewidth]{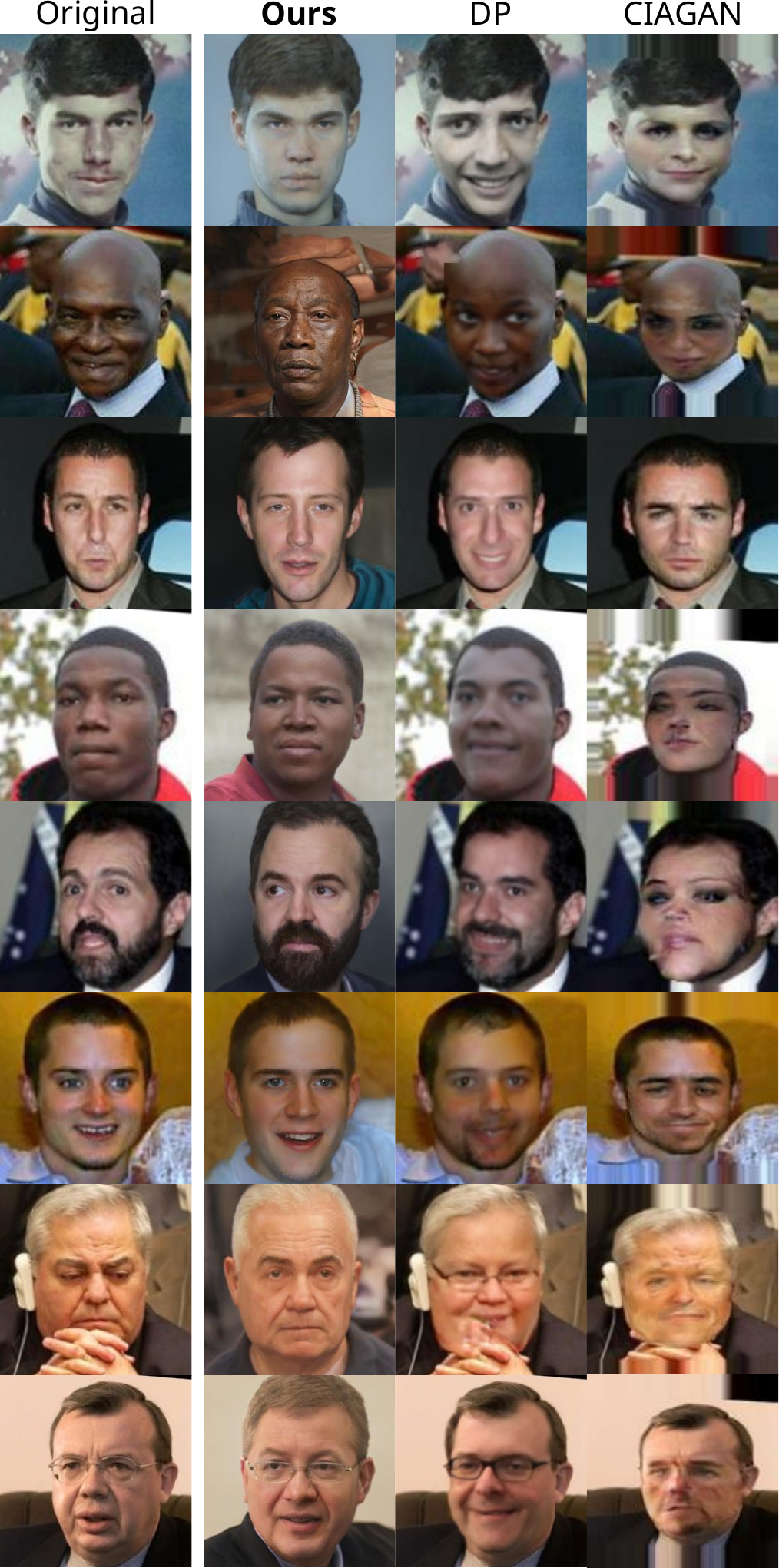}
                \caption{Anonymization results on LFW~\cite{LFWTech} in comparison to DeepPrivacy (DP)~\cite{hukkelas2019deepprivacy} and CIAGAN~\cite{maximov2020ciagan}.}
                \label{fig:lfw_qual}
            \end{figure}
    
        \subsubsection{Qualitative evaluation}\label{sec:qualitative}
            In this section, we make a qualitative comparison to our method and the SOTA. We show in Fig.~\ref{fig:celebahq_qual}, \ref{fig:lfw_qual} the original and anonymized images from the various methods on the two studied datasets. As can be clearly seen, our method is capable of retaining the facial attributes of the image to a much greater extent than the baselines -- \cite{hukkelas2019deepprivacy} often changes the expression and~\cite{maximov2020ciagan} often modifies the makeup of the images. Crucially, our method also succeeds in removing the background (and identifiable traces of the original image, such as particular clothing choices), which we have argued is of vital importance to true anonymization--removing any ability to infer the original image from the anonymized counterpart. More qualitative results can be found in the supplementary material~\ref{supp:additional_qual}.
    
    \subsection{Ablation study}\label{sec:ablation}
        In this section, we perform an ablation study on the value of the margin $m$ that controls the similarity between the identities. Concretely, we perform the same anonymization procedure for 50 epochs for the whole CelebA-HQ~\cite{liu2015faceattributes} dataset by changing only the value of $m$. In particular, we consider two extremes of $m=0.0$ and $m=0.9$. The larger the value of $m$, the more the resulting identity ought to be close to the original, hence re-identification results should be worse-off, while facial attributes should be better preserved.

        We see from the results in Tab.~\ref{tab:ablation-2} that $m$ indeed offers this trade-off, with the higher value of $m$ offering better attribute preservation at the cost to slightly worse identity re-identification performance. As one expects, we also see a lower value of FID using the higher margin, given the image is encouraged to be more close to the original.

\section{Conclusions}\label{sec:conclusions}
    In this paper, we presented a novel anonymization framework that directly optimizes the images' latent representation in the latent space of a pre-trained GAN, using a novel margin-based identity loss and an attribute preservation loss. Our method acts directly in the latent space of pre-trained GANs, avoiding the burden of the need to train complex networks. We showed that our method is capable of anonymizing the identity of the images whilst better-preserving the facial attributes, leading to better de-identification and facial attribute preservation than SOTA.


{\setlength{\parindent}{0.0cm}
\textbf{Acknowledgments:} This work was supported by the EU H2020 AI4Media No. 951911 project.
}

\clearpage
\appendix
\section{Supplementary Material}\label{supp}

    \subsection{Pre-trained attribute classifiers}\label{supp:attr_classifiers}
        As discussed in the Sect.~4.2.2 of the main paper, for the evaluation of the attribute preservation ability of the proposed and other state-of-the-art anonymization methods, in the case of the LFW~\cite{LFWTech} dataset, due to the absence of attribute labels, we obtain pseudo-labels provided by two pre-trained attribute classifiers, both pre-trained on the CelebA~\cite{liu2015faceattributes} dataset. Specifically, we use the pre-trained models provided by Anycost GAN~\cite{2021anycost} and Talk-to-Edit~\cite{2021talk2edit}. The former provides predictions (i.e., pseudo-labels) for the whole set of the 40 attributes of the CelebA dataset, while the latter for 5 of them (namely, \textit{``Bangs'', ``Eyeglasses'', ``Smiling'', ``No\_Beard''} and \textit{``Young''}). By using the aforementioned pseudo-labels, we performed the training/evluation process similarly to the case of the CelebA-HQ~\cite{liu2015faceattributes} dataset.

    \subsection{Additional qualitative results}\label{supp:additional_qual}
    
\begin{figure}[t]
            \centering
            \includegraphics[width=\linewidth]{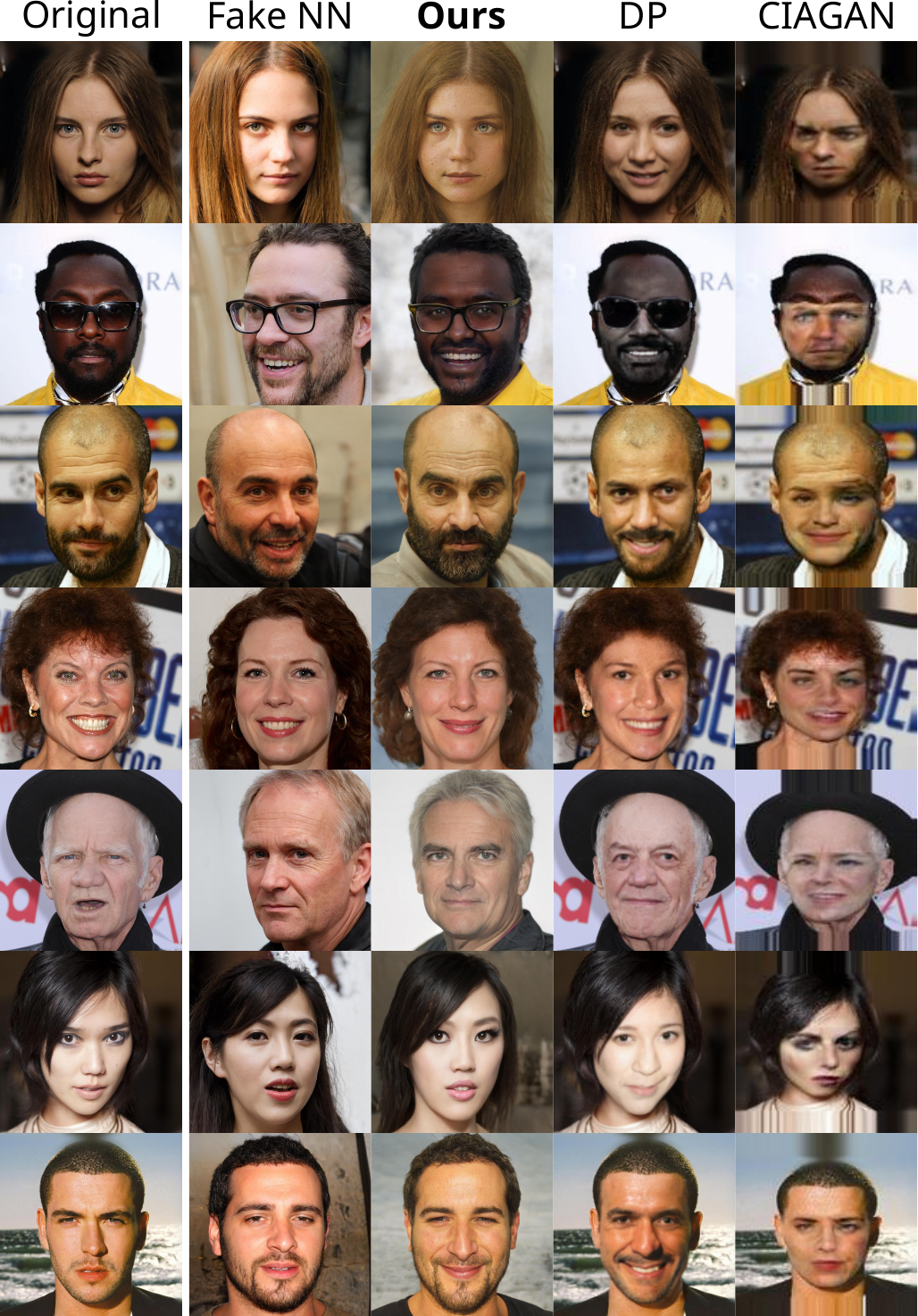}
            \caption{Anonymization results  of the proposed method in comparison to DeepPrivacy~\cite{hukkelas2019deepprivacy} and CIAGAN~\cite{maximov2020ciagan} on the CelebA-HQ~\cite{liu2015faceattributes} dataset. ``Fake NN'' denotes the nearest fake neighbor of each real image, obtained in the pre-trained FaRL~\cite{zheng2022farl} image representation space.}
            \label{fig:celebahq_qual_supp}
        \end{figure}

\begin{figure}[t]
            \centering
            \includegraphics[width=\linewidth]{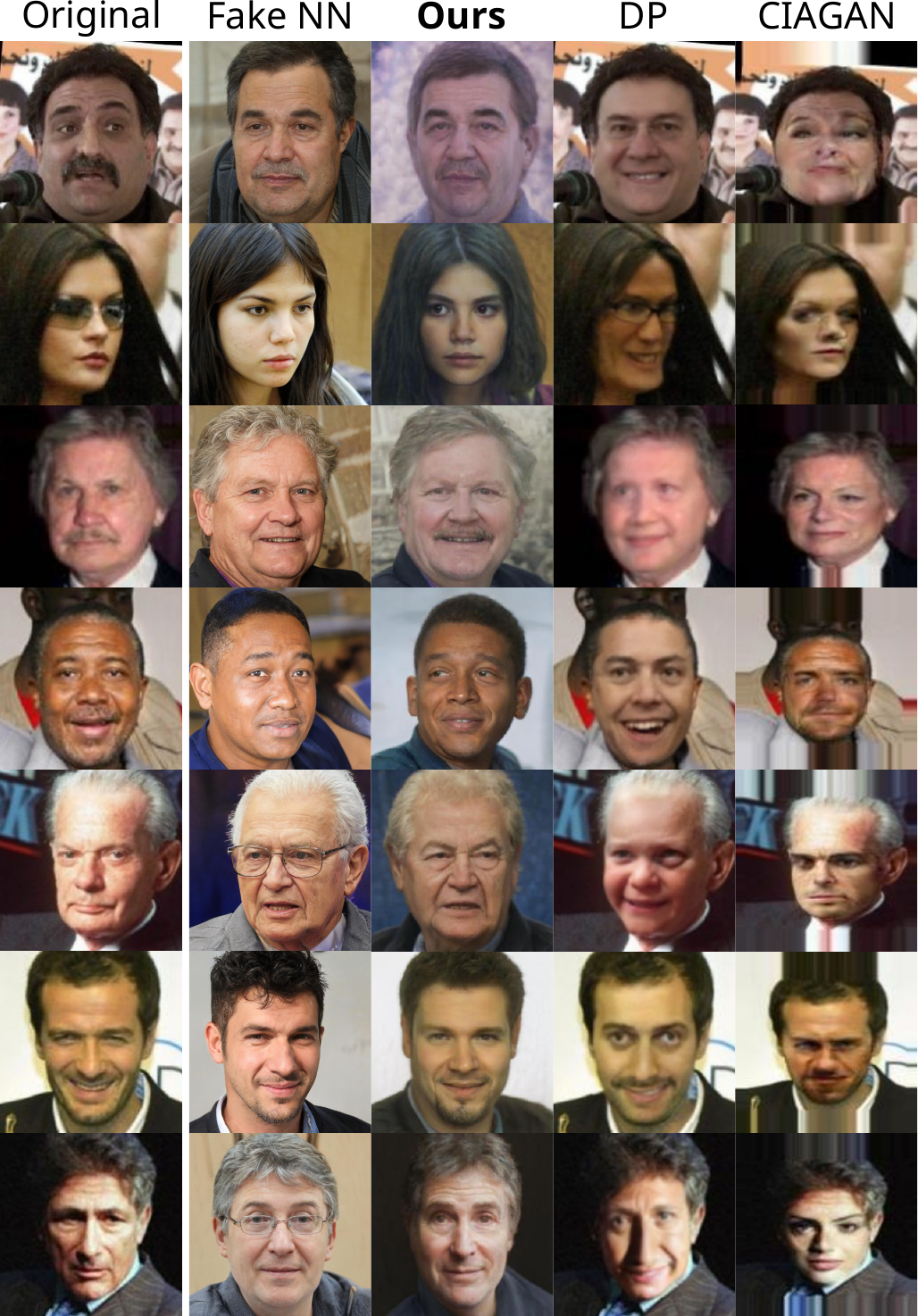}
            \caption{Anonymization results  of the proposed method in comparison to DeepPrivacy~\cite{hukkelas2019deepprivacy} and CIAGAN~\cite{maximov2020ciagan} on the LFW~\cite{LFWTech} dataset. ``Fake NN'' denotes the nearest fake neighbor of each real image, obtained in the pre-trained FaRL~\cite{zheng2022farl} image representation space.}
            \label{fig:lfw_qual_supp}
        \end{figure}

        In this section, we provide additional qualitative results of the proposed method in comparison to two state-of-the-art works, namely DeepPrivacy~\cite{hukkelas2019deepprivacy} and CIAGAN~\cite{maximov2020ciagan}, in both the CelebA-HQ~\cite{liu2015faceattributes} and the LFW~\cite{LFWTech} datasets, in Figs.~\ref{fig:celebahq_qual_supp},\ref{fig:lfw_qual_supp}, respectively. We observe that, in both datasets, the proposed method arrives at anonymized versions of the real face images that preserve more effectively both a certain level of similarity with the real ones and certain attributes (such as the skin tone and overall texture, facial hair, etc). By contrast, the state-of-the-art works~\cite{hukkelas2019deepprivacy,maximov2020ciagan} either lead to poor image quality (CIAGAN~\cite{maximov2020ciagan}) or/and fail to preserve certain facial attributes (DeepPrivacy~\cite{hukkelas2019deepprivacy}). This is also shown quantitatively in the Sect.~4 of the main paper. Finally, similarly to the previous section, we report as ``Fake NN'' the fake nearest neighbor (in the pre-trained FaRL~\cite{zheng2022farl} space) of each real image and observe that the proposed method provides an intuitive yet very simple way of initializing the latent codes that are then optimized in order to generate the anonymized face images.
        
    \subsection{Insights in the optimization process}\label{supp:optim_proc}
        
\begin{figure}[t]
            \centering
            \includegraphics[width=0.95\linewidth]{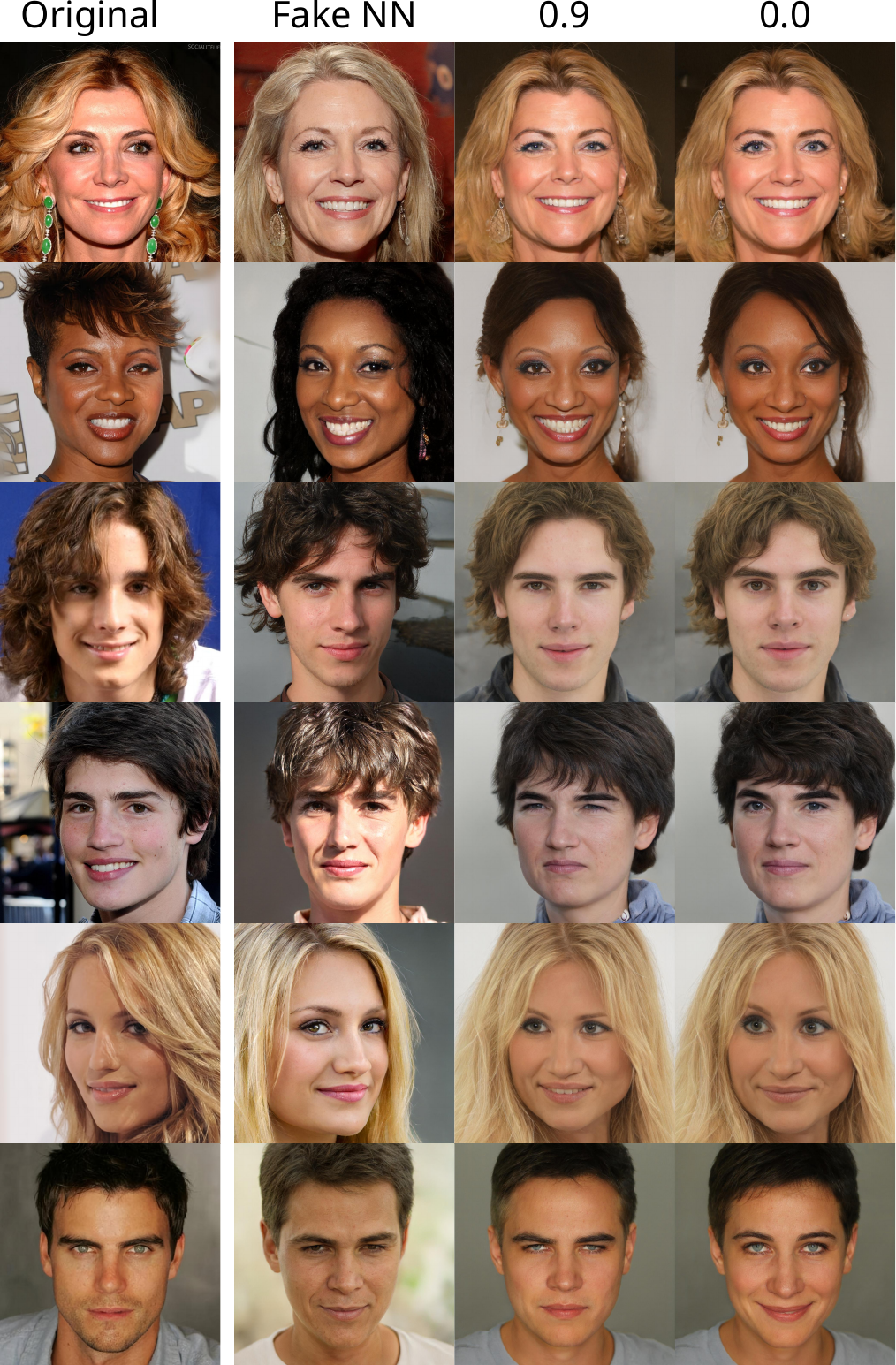}
            \caption{Anonymization results of the proposed method for $m\in\{0.0,0.9\}$ on the CelebA-HQ~\cite{liu2015faceattributes} dataset. ``Fake NN'' denotes the nearest fake neighbor of each real image, obtained in the FaRL~\cite{zheng2022farl} image representation space.}
            \label{fig:ablation_m}
        \end{figure}

        In this section, we provide additional insight on the two stages of the proposed framework, i.e., the pairing of real images with fake ones and the latent code optimization (discussed in detail in Sect.~3.2 and Sect.~3.3 of the main paper, respectively). In Fig.~\ref{fig:ablation_m} we show qualitative results of the proposed method for two values of the $m$ hyperparameter (introduced in Sect.~3.3 in the main paper) that controls the dissimilarity between the real and the anonymized face images. Specifically, when $m\to0$, the proposed identity loss (Eq.~(2) in the main paper) imposes orthogonality between the features of the real and the anonymized face images, leading to anonymized faces with large identity difference compared to the corresponding real ones. By contrast, when $m\to1$, the proposed identity loss imposes high similarity between the features of the real and the anonymized face images. Also, for each real image in Fig.~\ref{fig:ablation_m}, we report its corresponding fake nearest neighbor (obtained as described in detail in Sect.~3.2 in the main paper), denoted as ``Fake NN''. That is, the fakes image drawn from a pool of generated images that are closest to the real ones in the feature space of the pre-trained FaRL~\cite{zheng2022farl}. The latent codes of these fake neighbors are used for initializing the latent codes that are optimized for anonymizing the respective real images. As shown in Fig.~\ref{fig:ablation_m}, the fake nearest neighbor (``Fake NN'') provides a meaningful starting point for the optimization of the anonymized latent code, but does not limit the final anonymized generation with respect to facial attributes, the skin tone, or the head pose. Finally, we observe that $m=0.9$ leads to anonymized faces with higher identity similarity to the real ones compared to $m=0.0$.
    
    \subsection{Processing time}\label{supp:proc_time}
        
        As discussed in the main paper, the proposed framework incorporates only pre-trained networks (i.e., StyleGAN2's~\cite{stylegan2_karras20cvpr} generator $\mathcal{G}$, e4e~\cite{2021e4e}, FaRL's~\cite{zheng2022farl} ViT-based image encoder $\mathcal{E}_\mathcal{F}$, and ArcFace~\cite{Deng2021arcface} identity encoder $\mathcal{E}_\mathcal{A}$, as shown in Fig.~1 in the main paper), while at the same time the only trainable parameters are those of the latent codes that are optimized to anonymize the real images (i.e., $5\times512$ parameters per image). Learning each latent code requires $\sim3$ sec/epoch in 1 Nvidia RTX 3090 (we train for 50 epochs), while generating an anonymized image from its optimized latent code requires a single forward pass of the optimized latent code through $\mathcal{G}$ (0.05 sec).

    \subsection{Limitations}\label{supp:limitations}
        
        As discussed in the main paper (Sect.~3), the proposed framework relies on a pre-trained StyleGAN2~\cite{stylegan2_karras20cvpr} generator (typically pre-trained in the FFHQ~\cite{stylegan2_karras20cvpr} dataset) for generating the set of fake images (as described in Sect.~3.2 in the main paper), which are subsequently used for finding appropriate pairs (nearest fake neighbors in the FaRL~\cite{zheng2022farl} space) for each real image in order to initialize the latent codes ultimately optimized for the anonymization of the real images. This poses certain limitations to the proposed framework that reflect the limitations of the adopted GAN generator in generating faces statistically similar to the real ones, i.e., to the ones that will be anonymized. That is, the proposed method fails to anonymize real faces and to preserve all the relative attributes (e.g., hats) at the same time when the said attributes are not well-represented in the dataset that the adopted GAN generator has been trained with. Another limitation of the proposed framework concerns the inversion method that it incorporates (e.g., the e4e~\cite{2021e4e}), which might lead to unfaithful latent code inversions and thus affect the anonymization results.

{\small
\bibliographystyle{ieee_fullname}
\bibliography{egbib}
}

\end{document}